\documentclass{article}

\usepackage{microtype}
\usepackage{graphicx}
\usepackage{subcaption}
\usepackage{booktabs} 

\usepackage{hyperref}
\usepackage{tcolorbox}
\usepackage{longtable}
\usepackage{CJKutf8}
\usepackage{multirow}

\usepackage[preprint]{icml2026}

\usepackage{amsmath}
\usepackage{amssymb}
\usepackage{mathtools}
\usepackage{amsthm}
\usepackage{enumitem}

\usepackage[capitalize,noabbrev]{cleveref}

\theoremstyle{plain}

\theoremstyle{definition}

\theoremstyle{remark}

\usepackage[textsize=tiny]{todonotes}

\icmltitlerunning{A Colorization Framework for Aligning Cultural Semantics}

\begin{document} 

\twocolumn[
  \icmltitle{Reviving Ancient Paintings via Poem: A Colorization Framework for Aligning Cultural Semantics}



  \icmlsetsymbol{equal}{*}

  \begin{icmlauthorlist}
    \icmlauthor{Junming Gao}{zju}
    \icmlauthor{Biao Zhu}{hri}
    \icmlauthor{Xiaosong Wang}{zju}
    \icmlauthor{Tan Tang}{zju}
  \end{icmlauthorlist}

  \icmlaffiliation{zju}
  {Zhejiang University, Hangzhou, China}

  \icmlaffiliation{hri}
  {Hangzhou Research Institute of AI and Holographic Technology, Hangzhou, China}

  \icmlcorrespondingauthor{Tan Tang}{tan.tang@zju.edu.cn}

  \icmlkeywords{Ancient Painting, Diffusion, Image Colorization}

  \vskip 0.3in
]



\printAffiliationsAndNotice{}  

\begin{abstract}
The irreversible fading of ancient paintings disrupts the ``congruence between poems and paintings'', a core aesthetic principle where visual imagery harmonizes with literary inscriptions.
Although diffusion models provide strong generative priors, restoring
historically faithful colors remains difficult: visual restoration is
inherently ambiguous, while direct text guidance often causes modern
semantic bias, over-saturation, and cross-boundary color leakage.
To address this, we propose PoemColor, a poem-guided ancient painting colorization
framework.
Our method aligns poetic cultural semantics with painting restoration through
two key designs.
First, the Poetic Painting Projector (P3) converts implicit poetic context
into a classical color-aware condition via poem-to-palette pretraining,
reducing the ambiguity of poem-to-color mapping.
Second, Structure-Aware Semantic Attention (SASA) regulates how poetic color
semantics are injected into the diffusion backbone by jointly controlling
their propagation direction and regional injection strength.
In addition, we construct a hybrid restoration dataset that integrates
synthetic degradation with expert-restored artifacts, providing both scalable
supervision and real classical color references.
Extensive experiments demonstrate that our framework significantly outperforms state-of-the-art methods, delivering controllable colorization that revives both historical authenticity and poetic semantics.
\end{abstract}

\section{Introduction}

Ancient Chinese paintings embody the cross-modal heritage of
``congruence between poems and paintings''~\cite{tang2023pcolorizor}.
While ink structures are often preserved, colors have irreversibly faded
due to pigment aging and environmental exposure.
Existing chemical treatments~\cite{Ruiz2021a,Haslam2020a}
and digital scanning~\cite{BWei2003a,Chen2013a}
cannot recover such non-linear color degradation.
Therefore, restoring ancient paintings as a purely visual inverse problem
is highly ambiguous: the same faded trace may correspond to multiple
historically plausible colors, 
collapsing pixel-level mapping into unconstrained color hallucination.

Recent domain-specific frameworks have explored digital heritage
restoration~\cite{zhang2026traditional,hu2024sgrgan,yang2026digital},
and diffusion-based colorization methods have advanced natural image
colorization~\cite{song2020denoising,saharia2022palette,liang2025control}.
However, directly applying these paradigms to ancient paintings remains
problematic.
Purely visual methods mainly depend on residual degraded colors, which are
often unreliable in severely faded regions.
Meanwhile, general text-image models are dominated by modern corpora
~\cite{radford2021learning,tschannen2025siglip};
standard text guidance may introduce modern semantic bias and produce
over-saturated ``digital art'' colors inconsistent with classical aesthetics.
This motivates a culturally grounded paradigm:
using classical poetry as an external prior for painting restoration.
Since poetry and painting share a unified classical color system,
poetic context can provide a compact aesthetic blueprint to narrow the
ambiguous color hypothesis space.

However, translating implicit poetic context into pixel-level colors
introduces two challenges:
\begin{itemize}[leftmargin=12pt, topsep=2pt, itemsep=0pt]
    \item \textbf{Poem-to-Color Semantic Gap (C1):}
    Classical poems encode colors through symbolic, seasonal, and emotional
    expressions rather than explicit color descriptions.
    Directly injecting open-domain text embeddings into diffusion models
    activates an unstable color space, failing to adaptively converge to the
    restrained color distribution of ancient paintings.

    \item \textbf{Structure-Aware Color Propagation (C2):}
    Ancient paintings are organized by sparse ink contours and brushstroke
    boundaries.
    Conventional cross-attention or spatial conditioning mainly optimizes
    semantic correspondence but lacks explicit structural regulation,
    causing poem-induced colors to leak across weak boundaries and disrupt
    the original ink topology.
\end{itemize}

In this paper, we propose \textbf{PoemColor}, a plug-in adapter for ControlNet to perform
poem-guided ancient painting colorization.
To bridge C1, we introduce a \textbf{Poetic Painting Projector (P3)},
which maps implicit poetic semantics into a classical color
manifold represented by compact learnable color bases.
Unlike generic text adapters or visual color-query mechanisms, P3 does not
directly inject unconstrained text embeddings; instead, it projects poetic
context into a historically constrained palette subspace.
Moreover, structure-color decoupled latent anchoring is incorporated into
P3, where the native structure latent preserves ink structure and the
low-frequency color anchor stabilizes convergence toward classical color
distributions.

To address C2, we design \textbf{Structure-Aware Semantic Attention (SASA)}
to regulate where and how poetic colors are injected accurately.
SASA estimates boundary-aware structural guidance from the faded painting,
suppresses semantic feature components that tend to propagate across
boundary-normal directions, and adaptively modulates colorization intensity
with a dynamic spatial gate.
Therefore, P3 determines the historically plausible color subspace, while
SASA constrains its structure-consistent propagation path.
Crucially, data scarcity remains a fundamental bottleneck in existing
ancient painting restoration paradigms, as available paired data are often
extremely limited due to the rarity of expert-restored artifacts
~\cite{zhang2026traditional,hu2024sgrgan,yang2026digital}.
To overcome this limitation, we further construct a hybrid dataset by
integrating synthetic degradation with expert-restored artifacts, providing
both scalable supervision and real restoration references.

Extensive experiments show that our method outperforms SOTA baselines,
achieving the best LPIPS (0.334) and FID (19.84), while reducing color
distribution error by approximately 54\% (about 1.84).
These results indicate that poetic manifold projection, coupled
with structure-aware semantic propagation, provides a principled path toward
historically faithful ancient painting colorization.

\section{Related Work}

\subsection{Text-Guided Image Generation}
\label{sec:rel_semantic_prior}
Text-to-Image (T2I) generation has evolved from early Generative
Adversarial Networks~\cite{zhang2017stackgan,xu2018attngan}
to dominant Diffusion Models~\cite{rombach2022high,ramesh2022hierarchical},
which provide stronger fidelity and semantic controllability.
To reduce the uncertainty of pure text guidance, recent multimodal
frameworks introduce additional controls.
Structure-aware methods, including ControlNet-based approaches
~\cite{zhang2023adding,peng2024controlnext,yang2024pixel},
T2I-Adapter~\cite{mou2024t2i}, and OminiControl~\cite{tan2025ominicontrol},
inject spatial conditions such as edges or depth maps.
Appearance-aware methods, such as IP-Adapter~\cite{ye2023ip},
Uni-ControlNet~\cite{zhao2023uni}, and Composer~\cite{huang2023composer},
enable reference-guided style or attribute transfer.
Despite their success in general-purpose generation, these methods mainly
learn semantic priors from modern image-text corpora, making them less
suitable for ancient paintings whose colors follow restrained historical
palettes and classical aesthetic conventions.

Recent Poetry-to-Image frameworks
~\cite{li2021paint4poem,ijcai2022p696,peng2024fhs,peng2025poe2clp,
jamil2025poetry} demonstrate that classical verses can be translated into
visually coherent compositions.
However, they primarily focus on object-level or scene-level ``Form'',
such as aligning poetic entities with mountains, rivers, or pavilions,
while the historically grounded ``Color'' dimension remains less explored.
Moreover, recent studies on classical Chinese prompt optimization indicate
that modern language models can understand classical Chinese expressions,
but their alignment behavior under concise and obscure classical language
is still imperfect~\cite{huang2026obscure}.
This suggests that the key difficulty in our task is not merely parsing
poetry, but aligning abstract poetic semantics with the target color space.
Therefore, unlike generic semantic adapters or color-query mechanisms,
our Poetic Painting Projector (P3) explicitly projects poetic semantics
into a classical color manifold to reduce poem-to-color ambiguity.

\subsection{Controllable Image Colorization}
\label{sec:rel_structure_color}
Image colorization has been widely studied under automatic and interactive
settings.
GAN-based methods~\cite{zhu2017unpaired,isola2017image,kim2022bigcolor}
and Transformer-based methods
~\cite{weng2022ct,ji2022colorformer,huang2022unicolor,tang2025previvor}
can generate plausible colors from grayscale or degraded inputs.
Diffusion-based methods~\cite{zabari2023diffusing} further improve
colorization quality through strong generative priors, and recent controlled
diffusion models~\cite{zhang2023adding,liang2025control,
choi2025finecontrolnet} allow text prompts, sketches, strokes, or other
conditions to guide colorization.
Nevertheless, these methods are mostly designed for natural images or
general restoration, where perceptual realism and semantic plausibility are
the primary goals.
For ancient painting restoration, historically faithful colorization also
requires preserving sparse ink contours and brushstroke topology.

Spatial control frameworks can effectively preserve global geometry.
ControlNet-based methods~\cite{zhang2023adding,peng2024controlnext,
yang2024pixel}, T2I-Adapter~\cite{mou2024t2i}, and related adapters lock
structures through Canny edges, line drawings, or depth maps.
Appearance-aware approaches~\cite{ye2023ip,zhao2023uni,huang2023composer}
further enable style or reference transfer.
However, these controls mainly determine the generated structure or global
appearance, rather than explicitly regulating how color semantics propagate
within ink boundaries.
As a result, poem-induced colors may leak across fragile contours or
overwrite originally uncolored regions.
To address this limitation, our Structure-Aware Semantic Attention (SASA)
uses boundary-aware guidance and dynamic spatial modulation to constrain
poem-conditioned color propagation along ink- and brushstroke-consistent
regions.

\begin{figure*}[t]
    \centering
    \includegraphics[width=0.99\textwidth,height=0.5\textheight,keepaspectratio]{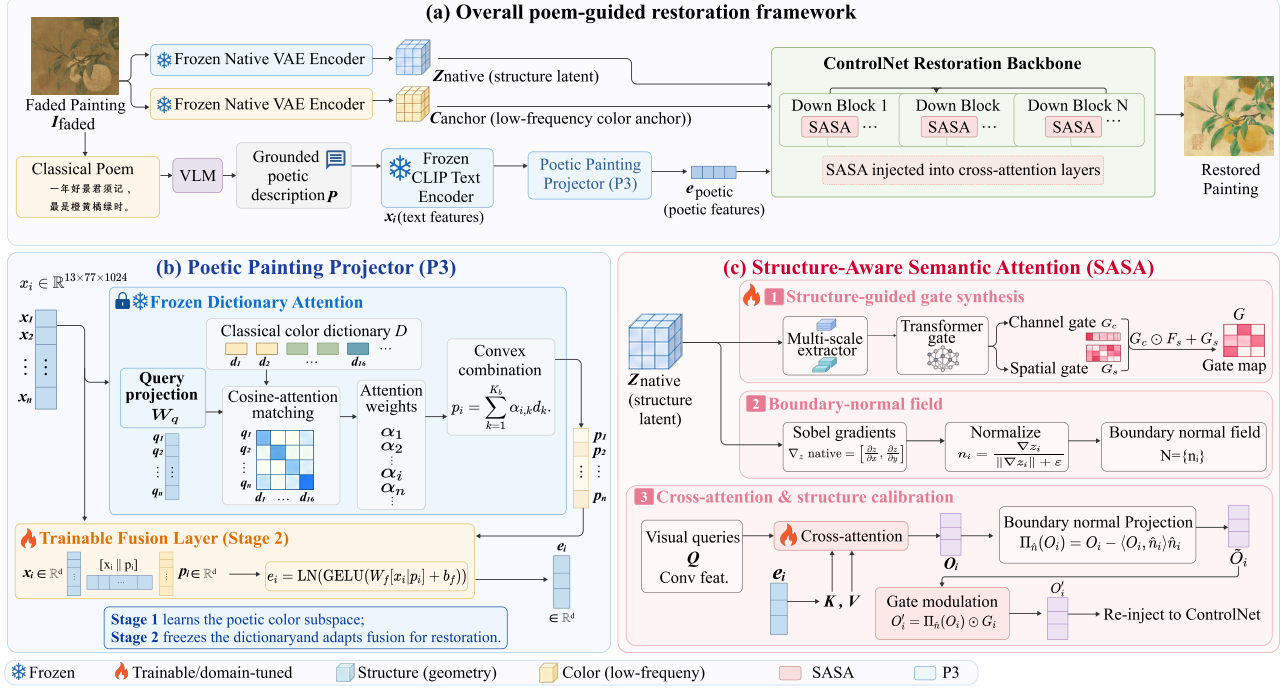}
    \caption{
    \textbf{Overall architecture of our poem-guided painting colorization framework.}
    \textbf{(a)} Given a faded painting $I_\mathrm{faded}$ and poem $L$,
    the framework extracts the structure latent $z_\mathrm{native}$,
    low-frequency color anchor $c_\mathrm{anchor}$, and poetic condition
    $e_\mathrm{poetic}$, then injects them into the ControlNet restoration
    backbone.
    \textbf{(b)} P3 projects CLIP text tokens onto a learnable classical
    color dictionary and fuses the projected color representation with the
    original text features.
    \textbf{(c)} SASA estimates a structure-aware gate $G$ and boundary-normal
    field $N$ from $z_\mathrm{native}$ to regulate the direction and strength
    of poem-conditioned color propagation within ink-consistent regions.
    }
    \label{fig:architecture}
\end{figure*}
\section{Methodology}

\subsection{Preliminary: LDM and ControlNet}

Our framework is built upon Latent Diffusion Models (LDMs)
~\cite{rombach2022high} and ControlNet~\cite{zhang2023adding}.
Given an input image $I$, a pretrained VAE encoder maps it to a latent
representation $z_0=\mathcal{E}(I)$.
The forward diffusion process gradually perturbs $z_0$ into $z_t$ with
Gaussian noise.
Under the standard $\epsilon$-parameterization, a time-conditioned U-Net
$\epsilon_\theta$ is trained to predict the added noise $\epsilon$:
\begin{equation}
\mathcal{L}_\mathrm{LDM}
=
\mathbb{E}_{z_t,t,c_\mathrm{txt},\epsilon}
\left[
\left\|
\epsilon-\epsilon_\theta(z_t,t,c_\mathrm{txt})
\right\|_2^2
\right].
\end{equation}
ControlNet introduces an additional trainable branch to process spatial
conditions $c*\mathrm{img}$, and injects the resulting features into the
frozen diffusion U-Net through zero-convolution layers:
\begin{equation}
\mathcal{L}_\mathrm{Ctrl}
=
\mathbb{E}_{z_t,t,c_\mathrm{txt},c_\mathrm{img},\epsilon}
\left[
\left\|
\epsilon-\epsilon_\theta
(z_t,t,c_\mathrm{txt},c_\mathrm{img})
\right\|_2^2
\right].
\end{equation}
In our task, ControlNet is extended from a spatial controller to a
poem-guided restoration backbone, where poetic semantics and painting
structures are jointly used for historically faithful colorization.

\subsection{PoemColor Overview}

Given a faded ancient painting
$I_\mathrm{faded}\in\mathbb{R}^{H\times W\times3}$
and its associated classical poem $L$, our goal is to predict historically
faithful colors while preserving the existing ink structures and brushstrokes.
This task introduces two coupled challenges:
the Poem-to-Color Semantic Gap (C1) and Structure-Aware Color Propagation
(C2).

To address these challenges, we propose PoemColor that transforms
a standard ControlNet into a poem-guided and structure-aware colorization
model.
As illustrated in Figure~\ref{fig:architecture}, our adapter consists of
two key components: Poetic Painting Projector (P3) and Structure-Aware
Semantic Attention (SASA).

\textbf{1) Translating Poetry into Painting Color Conditions.}
Instead of directly injecting open-domain text embeddings into the diffusion
model, P3 projects poetic semantics into a compact classical color subspace
and produces the poetic condition $e_\mathrm{poetic}$.
P3 is first pretrained with a lightweight poem-to-palette proxy task and
then frozen as a stable poetic color prior.
It is further coupled with two latent anchors: a native structure latent
$z_\mathrm{native}$ for preserving ink topology and a low-frequency color
anchor $c_\mathrm{anchor}$ for stabilizing convergence toward classical
palette distributions.

\textbf{2) Constraining Poetic Color Propagation with Painting Structure.}
SASA regulates where and how the poetic condition is injected into the
ControlNet attention layers.
It estimates boundary-aware structural guidance from $z_\mathrm{native}$,
suppresses semantic feature components that tend to propagate across
boundary-normal directions, and modulates the remaining features with a
dynamic spatial gate.
Thus, P3 determines the historically plausible color subspace, while SASA
constrains the structure-consistent propagation path of poem-induced colors.

\subsection{Poetic Painting Projector (P3)}
\label{sec:p3}

To address C1, P3 converts abstract poetic semantics into a low-rank
classical color representation.
Unlike visual color-query methods that learn color tokens mainly from image
features, P3 learns a poem-to-color projection from palette supervision,
thereby constraining poetic semantics within a compact classical color
basis space.

Given a classical poem $L$, we first employ a VLM~\cite{bai2025qwen2}
to distill the abstract poetic content into a grounded poetic description
$\mathcal{P}$~\cite{jamil2025poetry}.
A frozen CLIP text encoder then extracts token embeddings
\begin{equation}
c_\mathrm{txt}
=
\{\mathbf{x}_i\}_{i=1}^{N_\mathrm{tok}}
\in\mathbb{R}^{N_\mathrm{tok}\times1024},
\end{equation}
where $N_\mathrm{tok}$ denotes the maximum token length.

P3 maintains a learnable classical color dictionary
$\mathcal{D}\in\mathbb{R}^{K_b\times1024}$, where $K_b=16$.
For each token $\mathbf{x}_i$, a query vector is first generated:
\begin{equation}
\mathbf{q}_i
=
W_q\mathbf{x}_i+\mathbf{b}_q .
\end{equation}
The query is matched with normalized color bases by cosine attention:
\begin{equation}
\alpha_{i,k}
=
\frac{
\exp(\tau\langle\hat{\mathbf{q}}_i,\hat{\mathbf{d}}_k\rangle)
}{
\sum_{j=1}^{K_b}
\exp(\tau\langle\hat{\mathbf{q}}_i,\hat{\mathbf{d}}_j\rangle)
},
\end{equation}
where $\tau$ is a learnable temperature.
The projected color representation is obtained as:
\begin{equation}
\mathbf{p}_i
=
\sum_{k=1}^{K_b}
\alpha_{i,k}\mathbf{d}_k .
\end{equation}
This operation represents continuous poetic semantics as a convex
combination of a finite set of color bases, providing a rank-$K_b$
approximation of the classical color manifold.
The original text token and the projected color representation are fused by:
\begin{equation}
\mathbf{e}_i
=
\mathrm{LN}
\left(
\mathrm{GELU}
\left(
W_f[\mathbf{x}_i\Vert\mathbf{p}_i]+\mathbf{b}_f
\right)
\right),
\end{equation}
where $[\cdot\Vert\cdot]$ denotes channel-wise concatenation.
The resulting sequence
$e_\mathrm{poetic}=\{\mathbf{e}_i\}_{i=1}^{N_\mathrm{tok}}$
is used as the poetic condition for diffusion restoration.

Although P3 provides a poetic color prior, ancient painting restoration
also requires reliable structural preservation.
Following recent restoration studies showing that raw image controls may
introduce undesirable color shifts by overfitting to degraded backgrounds
~\cite{lin2024diffbir}, we decouple reliable structure from corrupted
chromatic traces.
A frozen native VAE encoder extracts:
\begin{equation}
z_\mathrm{native}
=
\mathcal{E}_\mathrm{native}(I_\mathrm{faded}),
\end{equation}
which preserves ink contours and brushstroke topology.
The residual degraded colors are used only as supplementary structural hints
rather than fixed color targets~\cite{tang2025previvor}.
In parallel, a domain-tuned VAE encoder extracts a low-frequency color anchor:
\begin{equation}
c_\mathrm{anchor}
=
\mathcal{E}_\mathrm{anchor}(I_\mathrm{faded}),
\end{equation}
which provides a coarse reference for the classical color distribution.
Together, $e_\mathrm{poetic}$, $z_\mathrm{native}$, and $c_\mathrm{anchor}$
form the P3 prior for the subsequent restoration stage.

\subsection{Structure-Aware Semantic Attention (SASA)}
\label{sec:sasa}

To address C2, we introduce Structure-Aware Semantic Attention (SASA) to
regulate the propagation of poetic color semantics inside ControlNet.
Existing spatial controllers mainly preserve global geometry through edges
or depth maps, but they do not explicitly control how text-induced color
features propagate with respect to ink boundaries.
SASA therefore uses $z_\mathrm{native}$ to estimate structural guidance and
constrains the attention output accordingly.
Compared with a pure edge mask or spatial gate, SASA jointly controls the
direction and magnitude of poetic color propagation: boundary-normal
projection suppresses cross-boundary semantic components, while the
structure-aware gate modulates the injection strength across regions.

\textbf{Structure-Guided Gate Synthesis.}
Given
$z_\mathrm{native}\in\mathbb{R}^{B\times4\times64\times64}$,
we extract multi-scale structural features using parallel depthwise
convolutions with different dilation scales:
\begin{equation}
U_s
=
\mathrm{UP}
\left(
\mathrm{DWConv}^{(s)}(z_\mathrm{native})
\right),
\quad
s\in\{1,2,4,8\},
\end{equation}
where $\mathrm{UP}(\cdot)$ denotes bilinear upsampling.
The multi-scale features are then concatenated with an explicit Sobel edge
map and fused by a $1\times1$ convolution:
\begin{equation}
F_\mathrm{str}
=
\mathrm{Conv}_{1\times1}
\left(
U_1\Vert U_2\Vert U_4\Vert U_8\Vert M_\mathrm{sobel}
\right).
\end{equation}
A lightweight self-attention block captures non-local structural
dependencies, and the resulting feature is decoded into a structure-aware
gate:
\begin{equation}
G
=
\sigma
\left(
\mathcal{F}_\mathrm{gate}
\left(
\mathrm{SA}(F_\mathrm{str})
\right)
\right)
\in\mathbb{R}^{B\times1\times64\times64},
\end{equation}
where $\sigma(\cdot)$ denotes the sigmoid function.
This gate estimates the spatial intensity of poem-conditioned color
injection.

\textbf{Boundary-Normal Feature Projection.}
SASA also estimates a boundary-normal field from the structure latent.
We compute grayscale gradients $(g_x,g_y)$ using a frozen Sobel operator and
normalize them as:
\begin{equation}
N
=
\frac{(g_x,g_y)}
{\sqrt{g_x^2+g_y^2+\epsilon}} .
\end{equation}
For each selected ControlNet attention layer, both $G$ and $N$ are
interpolated to the current feature resolution.
Let
$O\in\mathbb{R}^{B\times hw\times C}$
denote the standard cross-attention output between visual queries and
$e_\mathrm{poetic}$.
The 2D normal field is lifted to the feature dimension by a lightweight
linear projection and normalized at each spatial token, yielding $\hat{n}$.
For each spatial token $i$, SASA removes the feature component aligned with
the local boundary-normal direction:
\begin{equation}
\Pi_{\hat{n}}(O_i)
=
O_i
-
\langle O_i,\hat{n}_i\rangle\hat{n}_i .
\end{equation}
This projection suppresses cross-boundary semantic components and provides
direction-aware control over poetic color propagation.

\textbf{Structure-Aware Semantic Modulation.}
The remaining semantic features are modulated by the structure-aware gate:
\begin{equation}
O_i^\prime
=
\Pi_{\hat{n}}(O_i)\odot G_i .
\end{equation}
Here, the projection term controls the propagation direction, while the gate
$G_i$ controls the region-adaptive injection magnitude.
The refined output $O^\prime$ is injected back into the corresponding
attention layer.
In practice, SASA is applied to the input blocks and middle block of
ControlNet, where semantic conditions interact strongly with spatial
features.
This design encourages poem-conditioned colors to propagate within
ink- and brushstroke-consistent regions while reducing cross-boundary
color leakage.

\subsection{Optimization Objectives}
\label{sec:objectives}

Our optimization contains two stages: P3 proxy pretraining and diffusion
restoration training.

\textbf{P3 Pretraining Loss.}
Before training the restoration model, P3 is pretrained with a
poem-to-palette proxy task.
For each ancient painting, we extract $K_p=8$ dominant HSV colors by
K-Means and represent the sorted palette as:
\begin{equation}
\mathbf{c}
=
[h_1,s_1,v_1,\ldots,h_{K_p},s_{K_p},v_{K_p}]
\in\mathbb{R}^{3K_p}.
\end{equation}
Given the predicted palette $\hat{\mathbf{c}}$ and target palette
$\mathbf{c}$, the palette regression loss is:
\begin{equation}
\mathcal{L}_\mathrm{palette}
=
w_h\mathcal{L}_\mathrm{hue}
+
\mathcal{L}_\mathrm{sv},
\end{equation}
where
$\mathcal{L}_\mathrm{hue}=1-\cos(2\pi(\hat{h}-h))$
handles the circularity of hue, and $\mathcal{L}_\mathrm{sv}$ denotes the
MSE loss on saturation and value.
To avoid color-basis collapse, we further apply an orthogonality loss
$\mathcal{L}_\mathrm{ortho}$ on the normalized color dictionary and a
usage-balance loss $\mathcal{L}_\mathrm{usage}$ on the average assignment.
The P3 objective is:
\begin{equation}
\mathcal{L}_\mathrm{P3}
=
\mathcal{L}_\mathrm{palette}
+
\lambda_\mathrm{ortho}\mathcal{L}_\mathrm{ortho}
+
\lambda_\mathrm{usage}\mathcal{L}_\mathrm{usage}.
\end{equation}
After pretraining, the palette prediction head is discarded, and P3 is
frozen as the poetic color prior.

\textbf{Structure-Weighted Denoising Loss.}
For diffusion restoration, we follow the standard
$\epsilon$-parameterization and predict the noise $\hat{\epsilon}$ from
the noisy latent.
To make optimization consistent with SASA, we use the structure-aware gate
$G$ to reweight the denoising loss:
\begin{equation}
\mathcal{L}_\mathrm{denoise}
=
\mathrm{mean}
\left(
\left\|
\hat{\epsilon}-\epsilon
\right\|_2^2
\odot
\frac{G}{\mathrm{mean}(G)}
\right).
\end{equation}
This weighting encourages the model to focus on regions with high structural
affinity while avoiding trivial changes in the global gradient scale.


\section{Experiments}

\subsection{Experimental Setup}

\textbf{Datasets.}
We construct a hybrid dataset based on the PRevivor repository
~\cite{tang2025previvor} to address the severe scarcity of paired ancient
painting restoration data.
Directly collecting paired degraded/restored samples from online museum
sources and existing works is extremely difficult, and the available scale is
insufficient to train diffusion-based colorization models.
Therefore, following the common practice in old photo restoration
~\cite{wan2020bringing}, we generate a large-scale synthetic dataset by
applying a stochastic degradation strategy inspired by PRevivor
~\cite{tang2025previvor} to well-preserved paintings.
This synthetic construction provides scalable supervision for learning
fading patterns, and its effectiveness has been validated in PRevivor as
well as by our experimental results for training poem-guided colorization.
In addition, we collaborate with professional art restorers to obtain a
high-quality set of real degraded artifacts with manual ground-truth
colorization.
This expert-colorized subset is randomly partitioned, with 70\% used for
fine-tuning and the remaining 30\% reserved exclusively for quantitative
testing.
Finally, a separate collection of unpaired museum artifacts is gathered from
real-world collections to evaluate cross-domain generalization through
qualitative analysis and user studies.

\textbf{Baselines.}
We benchmark our method against a comprehensive suite of 11 state-of-the-art methods that span three dominant generative paradigms:
(1) classic \textbf{GAN/CNN-based adversarial approaches}, including DeOldify~\cite{salmona2022deoldify}, CycleGAN~\cite{zhu2017unpaired}, and BigColor~\cite{kim2022bigcolor};
(2) \textbf{Transformer-based architectures} that leverage attention for long-range dependency modeling, such as ColorFormer~\cite{ji2022colorformer}, CT2~\cite{weng2022ct}, UniColor~\cite{huang2022unicolor}, and DDColor~\cite{kang2023ddcolor};
and (3) contemporary \textbf{Diffusion-based models} representing the current state-of-the-art, namely DiffusingColors~\cite{zabari2023diffusing}, ControlNet~\cite{zhang2023adding}, L-CAD~\cite{weng2023cad}, and Control-Color~\cite{liang2025control}.
To ensure a fair comparison, all baselines were retrained or fine-tuned on our constructed hybrid dataset using their official implementations until convergence.

\textbf{Implementation Details.}
Our framework is built upon Stable Diffusion v2.1.
The Native Structure Stream uses the frozen SD VAE encoder, while the Color Stream VAE is fine-tuned for $20000$ steps with a learning rate of $1e-5$.
The grounded prompts are extracted using QwenVL2.5.
During training, the Denoising U-Net remains frozen. We optimize the ControlNet-based side-network and the SASA module using the AdamW optimizer with a learning rate of $1e-5$ and a batch size of 32.
Training is conducted on an NVIDIA H200 GPU for $30000$ steps.

\textbf{Evaluation Metrics.}
We perform a quantitative evaluation in two settings.
For Paired Data (the reserved 30\% of expert-colorized artifacts), we employ four metrics: PSNR and SSIM~\cite{wang2004image} for pixel-level and structural fidelity, LPIPS~\cite{zhang2018unreasonable} for perceptual similarity, and $\Delta$Colorfulness~\cite{hasler2003measuring} to quantify chromatic deviation from manual ground truth.
For Unpaired Data (real-world artifacts), we assess generalization performance using FID~\cite{heusel2017gans} for distribution quality and $\Delta$Colorfulness (Wasserstein Distance) for historical color authenticity.

\begin{figure*}[ht]
    \centering
    \includegraphics[width=0.83\textwidth]{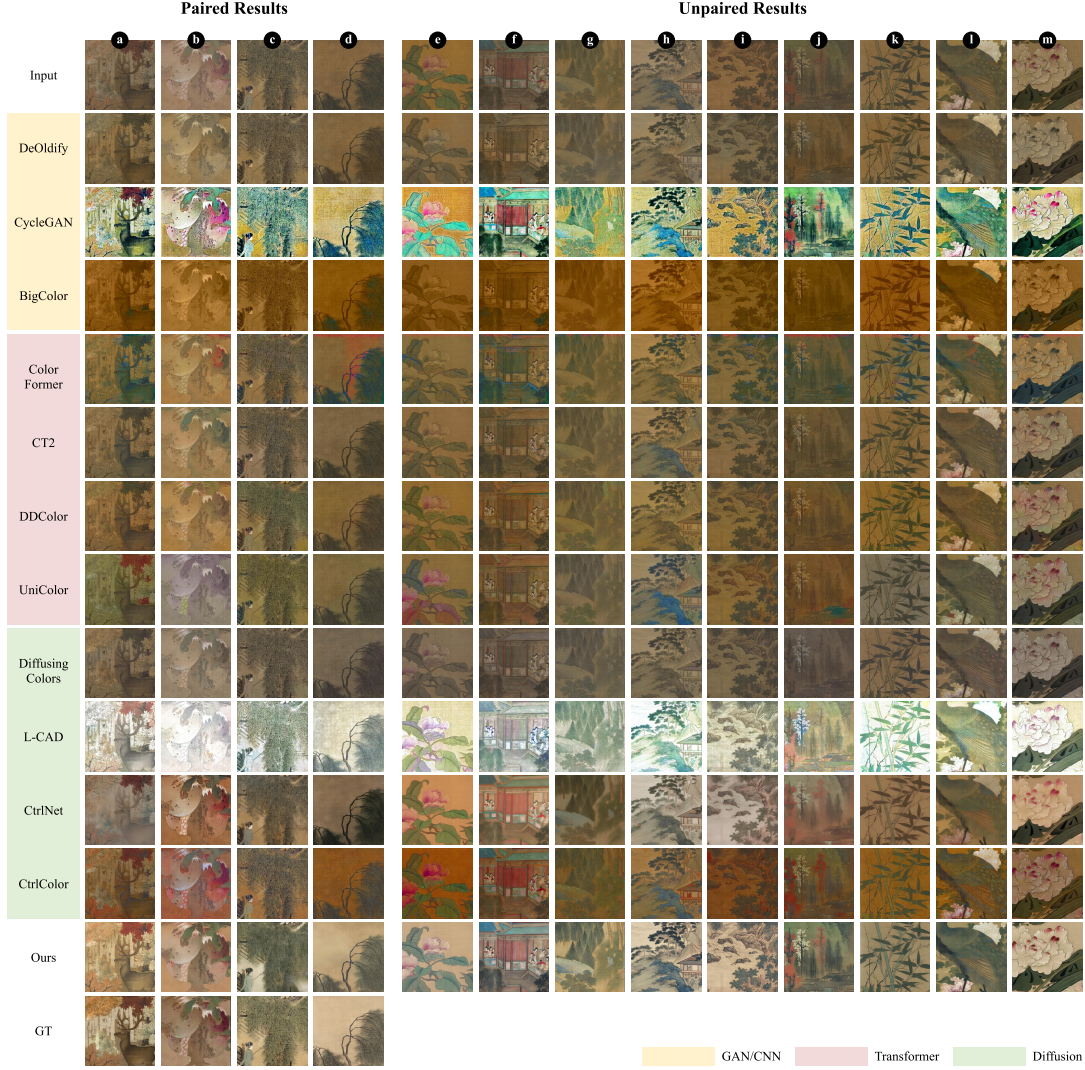} 
    \caption{\textbf{Qualitative comparison with state-of-the-art methods.} 
    We compare PoemColor against 11 baselines categorized by architecture: \colorbox{yellow!30}{GAN/CNN-based}, \colorbox{pink!30}{Transformer-based}, and \colorbox{green!20}{Diffusion-based} methods. 
    The left panel (\textbf{a-d}) displays reconstruction results on the Paired dataset with Ground Truth (GT), while the right panel (\textbf{e-m}) demonstrates colorization on Unpaired artifacts. 
    }
    \label{fig:visual_comp}
\end{figure*}

\subsection{Comparison}
\paragraph{Visual Comparison.}
Figure~\ref{fig:visual_comp} presents a qualitative comparison of challenging faded paintings in the real world, where we observe distinct failure modes in existing paradigms.
Legacy methods (e.g., DeOldify, ColorFormer, CT2) and BigColor exhibit a strong bias towards monochromatic tones, defaulting to conservative sepia, which indicates an overfitting to the global ``aged style''.
In terms of luminance, DDColor and UniColor fail to correct global dimming due to their rigid preservation of the input's luminance channel, resulting in visually dull outputs.
Conversely, lacking the semantic constraints of our Poetic Prior, CycleGAN and L-CAD tend to over-compensate, yielding results that are either excessively vibrant or severely over-exposed.
Furthermore, state-of-the-art diffusion baselines struggle with the domain gap: ControlColor hallucinates excessive saturation, incorrectly boosting the silk background (e.g, samples d and e) or generating neon-like artifacts (sample b), while standard ControlNet tends to smooth out high-frequency details, leading to blurred textures in intricate regions such as the peacock feathers in sample g.
In contrast, our framework successfully anchors the pigment distribution via the Color Stream and guides the atmosphere with the Poetic Prompt, recoloring historically authentic colors while preserving the underlying spatial structure.

\begin{table*}[!htbp]
    \centering
    \caption{\textbf{Quantitative comparisons with state-of-the-art methods.} We evaluate performance under two settings: \textbf{Paired Data} (compared against expert-colorized references) and \textbf{Unpaired Data} (assessing distributional alignment). Best results are highlighted in \textbf{bold}, and second-best results are \underline{underlined}.}
    \label{tab:quant_comparison}
    \resizebox{\textwidth}{!}{
    \begin{tabular}{llcccccc}
        \toprule
        \multirow{2}{*}{Type} & \multirow{2}{*}{Method} & \multicolumn{4}{c}{\textbf{Paired Data}} & \multicolumn{2}{c}{\textbf{Unpaired Data}} \\
        \cmidrule(lr){3-6} \cmidrule(lr){7-8}
         & & $\Delta$Colorfulness $\downarrow$ & LPIPS $\downarrow$ & SSIM $\uparrow$ & PSNR $\uparrow$ & FID $\downarrow$ & $\Delta$Colorfulness(WD) $\downarrow$ \\
        \midrule
        \multirow{3}{*}{GAN/CNN} 
         & DeOldify & \underline{9.44} & 0.431 & 0.591 & 11.94 & 40.73 & 7.15 \\
         & CycleGAN & 20.42 & 0.481 & 0.507 & 16.15 & 50.99 & 23.19 \\
         & BigColor & 12.56 & 0.678 & 0.476 & 9.04 & 45.63 & 5.53 \\
        \midrule
        \multirow{4}{*}{Transformer} 
         & ColorFormer & 12.02 & 0.491 & 0.573 & 11.64 & 38.95 & 4.75 \\
         & CT2 & 9.59 & 0.413 & 0.595 & 11.76 & 31.59 & 6.19 \\
         & DDColor & 10.26 & 0.455 & 0.579 & 11.64 & 33.89 & \underline{3.97} \\
         & UniColor & 13.30 & 0.445 & 0.590 & 11.97 & \underline{28.02} & 9.06 \\
        \midrule
        \multirow{4}{*}{Diffusion} 
         & DiffusingColors & 17.13 & 0.431 & 0.626 & 12.31 & 32.18 & 15.28 \\
         & L-CAD & 10.92 & 0.411 & \underline{0.670} & \underline{16.77} & 44.60 & 9.73 \\
         & ControlNet & 9.66 & \underline{0.404} & 0.573 & 16.41 & 35.10 & 5.11 \\
         & ControlColor & 13.34 & 0.570 & 0.467 & 10.94 & 34.92 & 9.27 \\
        \midrule
        \textbf{Ours} & \textbf{PoemColor} & \textbf{7.61} & \textbf{0.334} & \textbf{0.681} & \textbf{17.44} & \textbf{19.84} & \textbf{1.84} \\

        \bottomrule
    \end{tabular}
    }
\end{table*}

\paragraph{Quantitative Analysis.}
The quantitative evaluation in Table~\ref{tab:quant_comparison} confirms our visual findings. 
Legacy models (e.g., DeOldify, BigColor) exhibit the poorest perceptual scores (FID $>40.0$, High LPIPS), statistically confirming their inability to synthesize authentic textures and colors.
Although Transformer-based methods (e.g, UniColor) achieve competitive perceptual quality (FID $=$ 28.02), their low reconstruction fidelity (PSNR $\approx$ 11.6, SSIM $< 0.60$) quantifies the observed structural inconsistencies and color bleeding.
Among Diffusion-based baselines, although ControlNet and L-CAD (SSIM $=$ 0.670) maintain strong structural integrity, their lack of semantic guidance results in significant chromatic deviation (Unpaired $\Delta$Colorfulness $> 5.0$).
Our PoemColor framework achieves the best trade-off in all dimensions, dominating with the highest fidelity (PSNR 17.44, SSIM 0.681) and the most accurate pigment recovery ($\Delta$Colorfulness 1.84). This metric superiority confirms that our Structure-Gated Attention effectively translates poetic cues into precise visual features without compromising the underlying geometries.

\subsection{Ablation Studies}
\label{sec:ablation}

We conduct ablation studies to analyze the contribution of the key designs
in our framework.
Following common colorization evaluation protocols, we report FID for
perceptual distribution quality and Color WD for color-distribution
consistency.
All variants are evaluated under the same data split and inference setting,
and each ablation changes only the specified component while keeping the
remaining configuration unchanged.

\textbf{Effect of P3.}
Table~\ref{tab:ablation_all} studies the design of the Poetic Painting
Projector.
Directly using CLIP text embeddings yields limited color fidelity, showing
that open-domain text features are insufficient for poem-guided restoration.
Replacing P3 with a simple MLP adapter substantially reduces Color WD
(5.24 $\to$ 2.83), but its FID becomes slightly worse than direct CLIP,
indicating that generic text adaptation may improve color statistics without
fully preserving perceptual restoration quality.
Without poem-to-palette pretraining, P3 performs similarly to direct CLIP,
which confirms that the proxy task is necessary for learning a stable
poetic color prior.
Moreover, making the entire P3 trainable during restoration leads to clear
performance degradation, suggesting that diffusion gradients may corrupt the
pretrained color routing.
The full model achieves the best FID and Color WD, validating the
effectiveness of the proposed P3 design.

\textbf{Effect of SASA.}
Table~\ref{tab:ablation_all} evaluates the two key operations in SASA:
boundary-normal projection and structure-aware gate modulation.
Compared with the baseline without SASA, the projection-only variant improves
Color WD (6.76 $\to$ 4.44), but does not improve FID, suggesting that
direction-aware projection mainly regularizes color propagation rather than
directly enhancing global perceptual quality.
The gate-only variant improves both FID and Color WD, indicating that
region-adaptive modulation is crucial for activating poem-conditioned color
semantics.
The full model achieves the best results on both metrics, demonstrating that
direction-aware projection and region-adaptive gate modulation are
complementary for structure-consistent color propagation.

\textbf{Effect of structure-weighted denoising.}
Table~\ref{tab:ablation_all} shows the impact of the structure-weighted
denoising loss.
Removing this loss degrades both FID and Color WD, indicating that the gate
map is not only useful for forward semantic modulation in SASA, but also
provides an effective spatial weighting signal during diffusion optimization.
This consistency between forward attention regulation and backward denoising
supervision improves both perceptual quality and color-distribution fidelity.

\begin{table}[t]
    \centering
    \small
    \caption{
    \textbf{Ablation study of key components.}
    FID measures perceptual distribution quality, while Color WD denotes
    the Wasserstein distance of color distributions. Lower values are better.
    Each variant changes only the specified component from the full model.
    }
    \label{tab:ablation_all}
    \begin{tabular}{lcc}
    \toprule
    Model Variant & FID$\downarrow$ & Color WD$\downarrow$ \\
    \midrule
    \multicolumn{3}{l}{\textit{P3 design}} \\
    Direct CLIP       & 21.20 & 5.24 \\
    MLP Adapter       & 21.50 & 2.83 \\
    P3 w/o pretrain   & 21.25 & 5.19 \\
    P3 w/o freeze     & 23.19 & 5.47 \\
    \midrule
    \multicolumn{3}{l}{\textit{SASA components}} \\
    w/o SASA          & 22.32 & 6.76 \\
    Projection Only   & 22.49 & 4.44 \\
    Gate Only         & 21.94 & 2.52 \\
    \midrule
    \multicolumn{3}{l}{\textit{Structure-weighted objective}} \\
    w/o weighted loss & 22.44 & 3.08 \\
    \midrule
    \textbf{PoemColor} & \textbf{19.84} & \textbf{1.84} \\
    \bottomrule
    \end{tabular}
\end{table}

\subsection{User Evaluation}
\begin{figure}[h]
    \centering
    \includegraphics[width=0.40\textwidth]{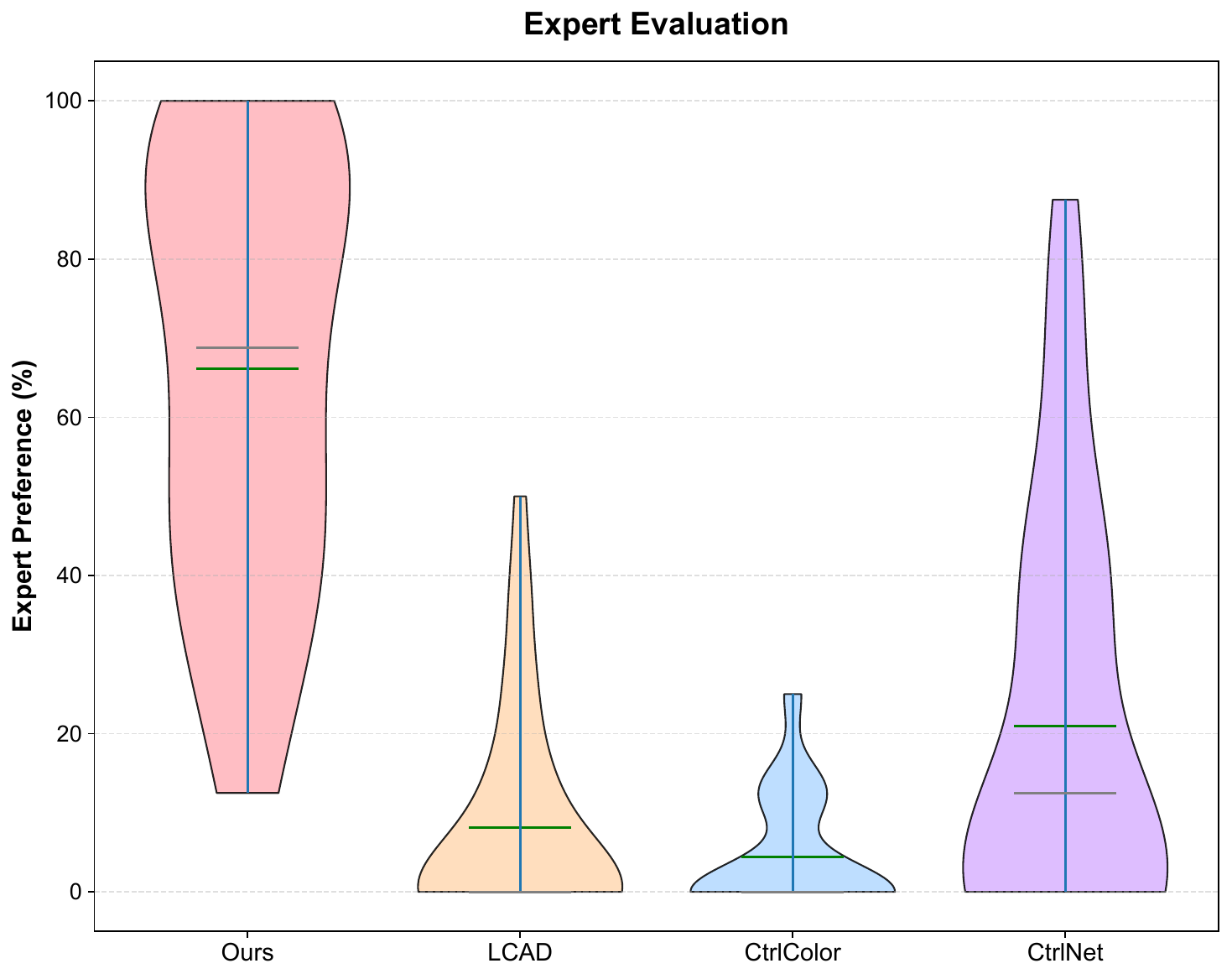}
    \caption{\textbf{Distribution of expert preference scores.} Our approach consistently stabilizes in the 65\%–70\% range, significantly surpassing top-tier baselines like L-CAD and ControlColor.}
    \label{fig:user_study}
\end{figure}

We conducted a blind user study with 8 academic experts, each with an average of 5 years of domain-specific experience, to compare our method against 11 baselines. For evaluation, 34 paintings were randomly selected and presented in randomized order. Experts assessed each result from three aspects: \textbf{Luminance} for visual harmony, \textbf{Structure} for brushstroke preservation, and \textbf{Color} for texture and chromatic alignment. Out of 272 total preference votes, our method received 179 votes (65.8\%), while the strongest baseline, ControlNet, received only 21.7\%. A Chi-Square Goodness-of-Fit test further confirms the statistical significance of this preference ($\chi^2=60.50$, $p<0.0001$), rejecting the null hypothesis of random choice with high confidence. Inter-rater agreement was also strong: for 14 out of 34 paintings, at least 7 of the 8 experts voted for our method. These results demonstrate that our method achieves the highest expert preference across all evaluation dimensions, as shown in Figure~\ref{fig:user_study}.

\section{Discussion and Conclusion}
\begin{figure}[htbp]
    \centering
    \includegraphics[width=0.49\textwidth,keepaspectratio]{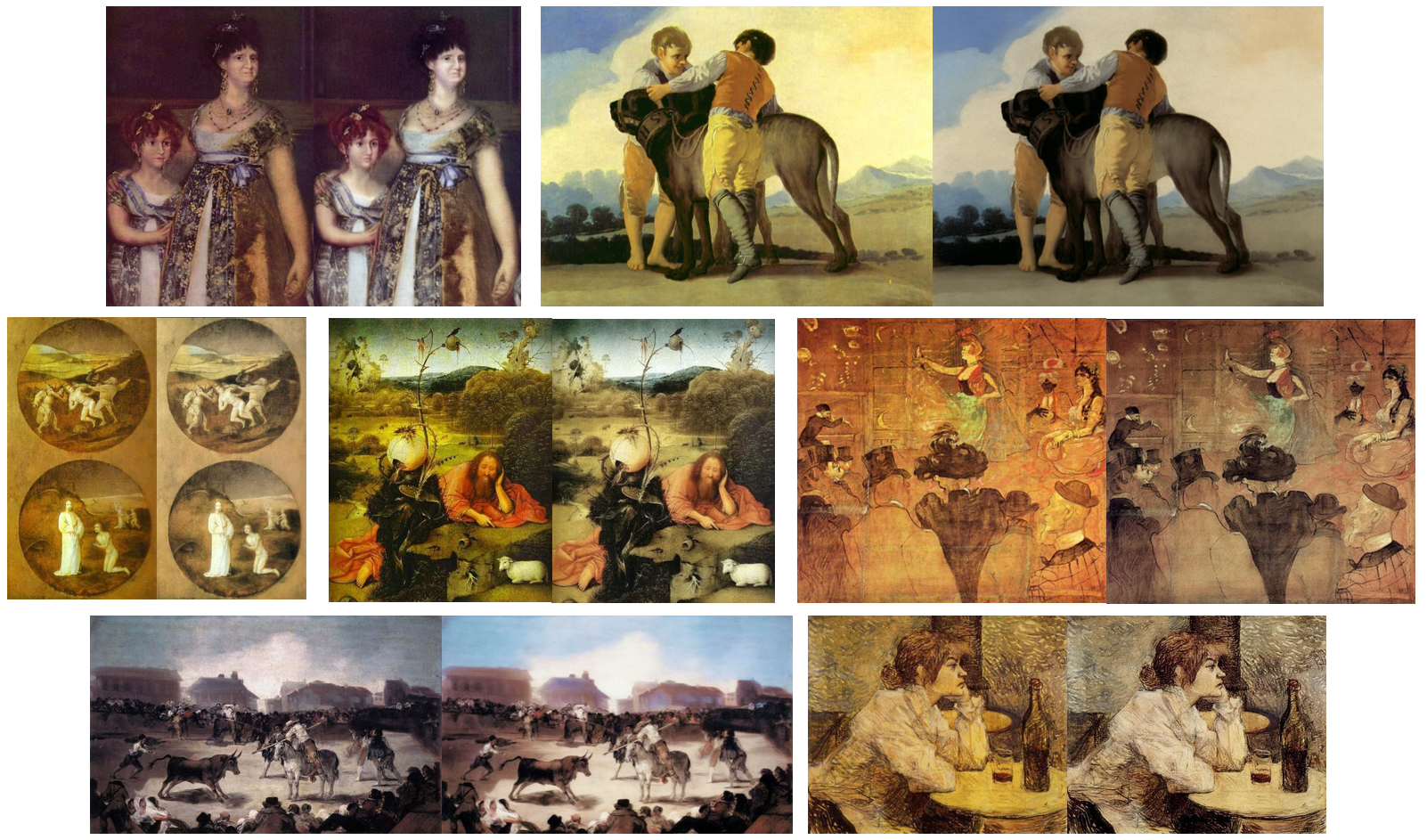}
    \caption{\textbf{Visual demonstration of cross-domain applicability on Western oil paintings.}
    The PoemColor successfully modulates color while preserving the original artwork’s distinct structural strokes.
    }
\label{fig:oil_vis}
\end{figure}

\paragraph{Generalization Discussion} 
We extended our evaluation to the 50-Artworks dataset~\cite{ikarus777_best_artworks}. 
Crucially, despite being trained exclusively in Chinese ancient paintings with zero exposure to Western paintings, PoemColor successfully decouples dense impasto structures from color dynamics without fine-tuning (Figure~\ref{fig:oil_vis}).
These results clearly demonstrate the universality of our Structure-Gated mechanism across diverse artistic modalities.

\paragraph{Limitations and Future Work} 
While our decoupled design ensures graceful degradation against individual modality failures (e.g., the Color Stream acting as a robust spectral anchor when VLM misinterprets metaphors), we acknowledge limitations in extreme scenarios.
First, spectral ambiguity between extremely faint pigments and substrate aging can lead to ``over-cleaning,'' where the VAE inadvertently filters out subtle background details.
Second, a ``Double-Blind'' scenario arises when severe degradation coincides with sparse poetic descriptions; in this information void, both visual and semantic priors fail, leading to conservative or desaturated predictions.
Future work will focus on fine-tuning VLMs on art-historical corpora and integrating hyperspectral data to resolve these ambiguities.

\paragraph{Conclusion}
In this paper, we propose PoemColor, a poem-guided colorization framework
for aligning cultural semantics with ancient painting restoration.
By translating implicit poetic context into classical color-aware conditions
through P3 and constraining their structure-consistent propagation with
SASA, PoemColor restores historically plausible colors while preserving
ink contours and brushstroke topology.
Together with our hybrid restoration dataset, this work demonstrates that
classical poetry can serve as an effective cultural prior for reducing color
ambiguity and advancing heritage painting restoration.

\newpage
\section*{Impact Statement}
This paper presents work whose goal is to advance the field of machine learning. There are many potential societal consequences of our work, none of which we feel must be specifically highlighted here.


\bibliography{icml26}
\bibliographystyle{icml2026}

\newpage


\end{document}